\PassOptionsToPackage{unicode}{hyperref}
\PassOptionsToPackage{hyphens}{url}
\PassOptionsToPackage{dvipsnames,svgnames,x11names}{xcolor}
\documentclass[
  letterpaper,
  DIV=11,
  numbers=noendperiod]{scrartcl}

\usepackage{amsmath,amssymb}
\usepackage{iftex}
\ifPDFTeX
  \usepackage[T1]{fontenc}
  \usepackage[utf8]{inputenc}
  \usepackage{textcomp} 
\else 
  \usepackage{unicode-math}
  \defaultfontfeatures{Scale=MatchLowercase}
  \defaultfontfeatures[\rmfamily]{Ligatures=TeX,Scale=1}
\fi
\usepackage{lmodern}
\ifPDFTeX\else  
\fi
\IfFileExists{upquote.sty}{\usepackage{upquote}}{}
\IfFileExists{microtype.sty}{
  \usepackage[]{microtype}
  \UseMicrotypeSet[protrusion]{basicmath} 
}{}
\makeatletter
\@ifundefined{KOMAClassName}{
  \IfFileExists{parskip.sty}{%
    \usepackage{parskip}
  }{
    \setlength{\parindent}{0pt}
    \setlength{\parskip}{6pt plus 2pt minus 1pt}}
}{
  \KOMAoptions{parskip=half}}
\makeatother
\usepackage{xcolor}
\ifx\paragraph\undefined\else
  \let\oldparagraph\paragraph
  \renewcommand{\paragraph}[1]{\oldparagraph{#1}\mbox{}}
\fi
\ifx\subparagraph\undefined\else
  \let\oldsubparagraph\subparagraph
  \renewcommand{\subparagraph}[1]{\oldsubparagraph{#1}\mbox{}}
\fi

\usepackage{color}
\usepackage{fancyvrb}

\DefineVerbatimEnvironment{Highlighting}{Verbatim}{commandchars=\\\{\}}
\usepackage{framed}
\definecolor{shadecolor}{RGB}{241,243,245}
\newenvironment{Shaded}{\begin{snugshade}}{\end{snugshade}}

\newcommand{\CommentTok}[1]{\textcolor[rgb]{0.37,0.37,0.37}{#1}}

\newcommand{\NormalTok}[1]{\textcolor[rgb]{0.00,0.23,0.31}{#1}}

\newcommand{\OtherTok}[1]{\textcolor[rgb]{0.00,0.23,0.31}{#1}}

\providecommand{\tightlist}{%
  \setlength{\itemsep}{0pt}\setlength{\parskip}{0pt}}\usepackage{longtable,booktabs,array}
\usepackage{calc} 
\usepackage{etoolbox}
\makeatletter
\patchcmd\longtable{\par}{\if@noskipsec\mbox{}\fi\par}{}{}
\makeatother
\IfFileExists{footnotehyper.sty}{\usepackage{footnotehyper}}{\usepackage{footnote}}
\makesavenoteenv{longtable}
\usepackage{graphicx}
\makeatletter
\def\maxwidth{\ifdim\Gin@nat@width>\linewidth\linewidth\else\Gin@nat@width\fi}
\def\maxheight{\ifdim\Gin@nat@height>\textheight\textheight\else\Gin@nat@height\fi}
\makeatother
\setkeys{Gin}{width=\maxwidth,height=\maxheight,keepaspectratio}
\makeatletter
\def\fps@figure{htbp}
\makeatother
\newlength{\cslhangindent}
\newlength{\csllabelwidth}
\newlength{\cslentryspacingunit} 
\newenvironment{CSLReferences}[2] 
 {
  \setlength{\parindent}{0pt}
  \ifodd #1
  \let\oldpar\par
  \def\par{\hangindent=\cslhangindent\oldpar}
  \fi
  \setlength{\parskip}{#2\cslentryspacingunit}
 }%
 {}
\usepackage{calc}

\usepackage{fvextra}
\DefineVerbatimEnvironment{Highlighting}{Verbatim}{breaklines,commandchars=\\\{\}}
\DefineVerbatimEnvironment{OutputCode}{Verbatim}{breaklines,commandchars=\\\{\}}
\KOMAoption{captions}{tableheading}
\makeatletter
\@ifpackageloaded{caption}{}{\usepackage{caption}}
\AtBeginDocument{%
\ifdefined\contentsname
  \renewcommand*\contentsname{Table of contents}
\else
  \newcommand\contentsname{Table of contents}
\fi
\ifdefined\listfigurename
  \renewcommand*\listfigurename{List of Figures}
\else
  \newcommand\listfigurename{List of Figures}
\fi
\ifdefined\listtablename
  \renewcommand*\listtablename{List of Tables}
\else
  \newcommand\listtablename{List of Tables}
\fi
\ifdefined\figurename
  \renewcommand*\figurename{Figure}
\else
  \newcommand\figurename{Figure}
\fi
\ifdefined\tablename
  \renewcommand*\tablename{Table}
\else
  \newcommand\tablename{Table}
\fi
}
\@ifpackageloaded{float}{}{\usepackage{float}}
\floatstyle{ruled}
\@ifundefined{c@chapter}{\newfloat{codelisting}{h}{lop}}{\newfloat{codelisting}{h}{lop}[chapter]}
\floatname{codelisting}{Listing}

\makeatother
\makeatletter
\@ifpackageloaded{caption}{}{\usepackage{caption}}
\@ifpackageloaded{subcaption}{}{\usepackage{subcaption}}
\makeatother
\makeatletter
\@ifpackageloaded{tcolorbox}{}{\usepackage[skins,breakable]{tcolorbox}}
\makeatother
\makeatletter
\@ifundefined{shadecolor}{\definecolor{shadecolor}{rgb}{.97, .97, .97}}
\makeatother
\ifLuaTeX
  \usepackage{selnolig}  
\fi
\IfFileExists{bookmark.sty}{\usepackage{bookmark}}{\usepackage{hyperref}}
\IfFileExists{xurl.sty}{\usepackage{xurl}}{} 
\hypersetup{
  pdfauthor={Gregor Betz (Logikon AI, KIT)},
  colorlinks=true,
  linkcolor={blue},
  filecolor={Maroon},
  citecolor={Blue},
  urlcolor={Blue},
  pdfcreator={LaTeX via pandoc}}

\title{\emph{Guided Reasoning}\\
A Non-Technical Introduction\\
{[}Logikon Version \texttt{v0.2.0}{]}}
\author{Gregor Betz (Logikon AI, KIT)}
\date{2024-01-08}

\begin{document}
\maketitle
\ifdefined\Shaded\renewenvironment{Shaded}{\begin{tcolorbox}[enhanced, boxrule=0pt, frame hidden, breakable, interior hidden, sharp corners, borderline west={3pt}{0pt}{shadecolor}]}{\end{tcolorbox}}\fi

\textbf{Abstract.} We introduce the concept and a default implementation
of \emph{Guided Reasoning}. A multi-agent system is a Guided Reasoning
system iff one agent (the guide) primarily interacts with other agents
in order to improve reasoning quality. We describe Logikon's default
implementation of Guided Reasoning in non-technical terms. This is a
living document we'll gradually enrich with more detailed information
and examples.

Code:
\href{https://github.com/logikon-ai/logikon}{github.com/logikon-ai/logikon}

Demo:
\href{https://huggingface.co/spaces/logikon/benjamin-chat}{huggingface.co/spaces/logikon/benjamin-chat}

\hypertarget{introduction}{%
\section{Introduction}\label{introduction}}

\begin{description}
\tightlist
\item[Definition Guided Reasoning (general).]
A multi-agent system that comprises a \emph{guide agent} and at least
one \emph{client agent} is a Guided Reasoning system iff the guide
systematically and primarily interacts with the clients in order to
elicit and shape client reasoning such that it complies with a given
\emph{method M}.
\end{description}

The reasoning method \emph{M} might be specified in the form of
standards and criteria, paradigmatic examples, or detailed rules and
instructions.

So, a coach that helps a business unit to carry out a SWOT analysis, a
kid assisting their granny to solve a crossword puzzle, or a Socratic
dialog (Nelson 2002) are \textbf{examples} of Guided Reasoning systems.
Rule-based consumer software can be part of a Guided Reasoning system,
too, for example when a small enterprise uses an accounting software to
set up its tax return, or to comply with financial regulation. Vice
versa, humans may figure as guides when supervising and steering
advanced GenAI systems (human in the loop).

The \textbf{prima facie case} for AI-AI Guided Reasoning rests on the
following assumptions:

\begin{enumerate}
\def\labelenumi{\arabic{enumi}.}
\tightlist
\item
  AI systems ought to give and explain correct answers.
\item
  AI systems can only faithfully explain their answers if the latter are
  based on explicit reasoning.
\item
  Poor reasoning undermines the ability of AI systems to give correct
  answers.
\item
  Strong domain experts are not necessarily able to follow advanced
  reasoning methods.
\end{enumerate}

In order to create explainable and accurate AI systems under these
assumptions, the principle of cognitive specialization suggests to build
extra AI experts for reasoning methods (meta-reasoning specialists),
which can work together with different domain experts. Guided Reasoning
is a promising design pattern for advanced GenAI apps because it allows
for effective division of cognitive labour.

This non-technical report presents Logikon's default implementation of
Guided Reasoning, where client agents, when facing a decision problem,
are steered towards exploring and systematically evaluating pro and con
arguments.

The report gives, in the next section, a
\protect\hyperlink{user-interactions-with-an-ai-guided-reasoning-system}{high-level
overview of how users may interact with a Guided Reasoning system},
subsequently unpacks
\protect\hyperlink{guiding-client-reasoners-in-balancing-pros-and-cons}{Logikon's
default implementation of Guided Reasoning} (balancing pros and cons),
and finally describes the
\protect\hyperlink{informal-argument-mapping-workflow}{AI argument
mapping workflow} that is part of the balancing process. Moreover, we
provide some pointers to \protect\hyperlink{related-work}{related work}.

\hypertarget{user-interactions-with-an-ai-guided-reasoning-system}{%
\section{User Interactions with an AI Guided Reasoning
System}\label{user-interactions-with-an-ai-guided-reasoning-system}}

\begin{figure}

{\centering 

\begin{figure}[H]

{\centering \includegraphics[width=6.5in,height=6.63in]{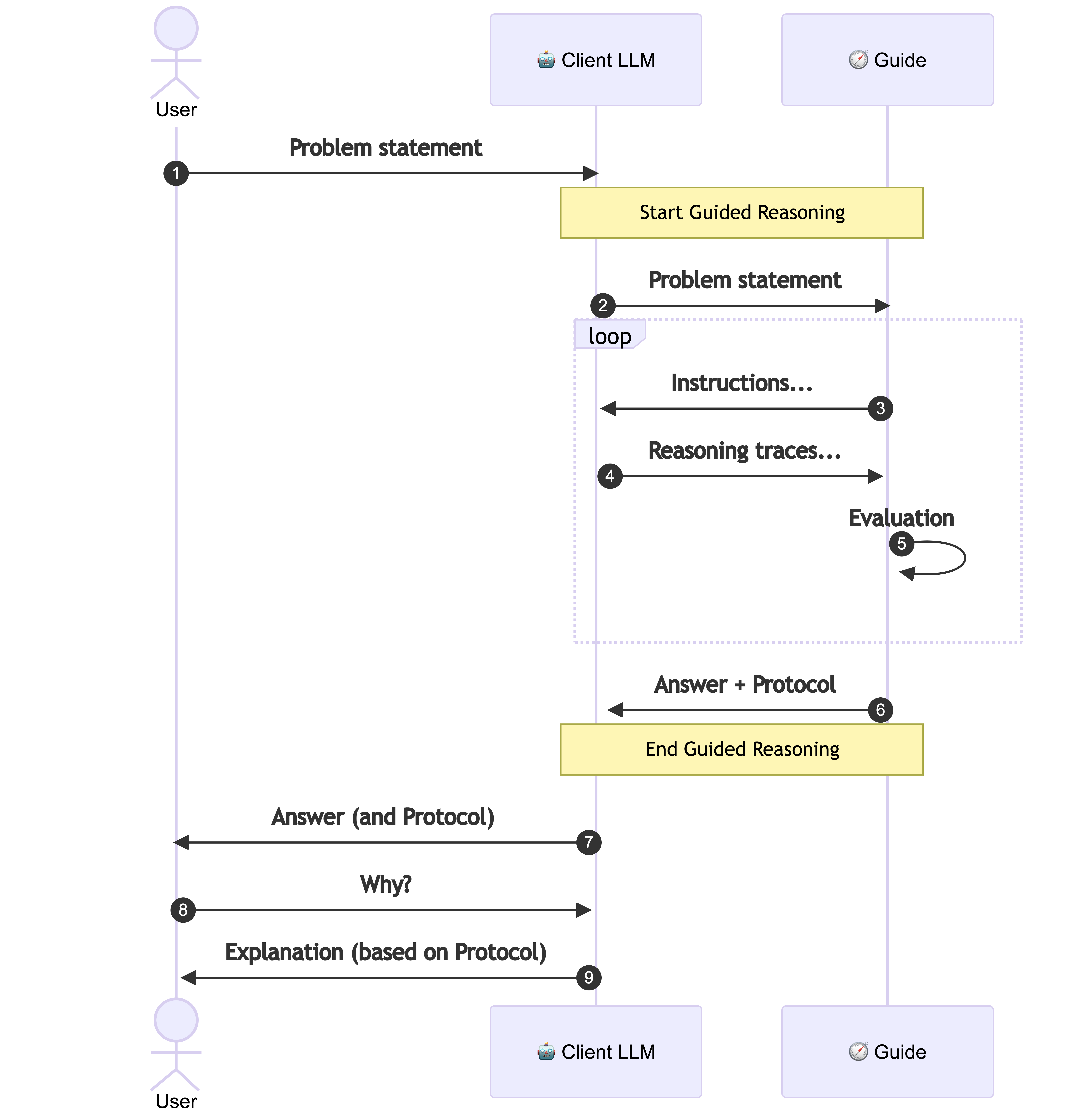}

}

\end{figure}

}

\caption{\label{fig-global-gr}User interactions with a Guided Reasoning
systems.}

\end{figure}

Figure~\ref{fig-global-gr} shows how users may interact with a Guided
Reasoning system, and sketches the interactions between client and guide
within that system. Let's walk through these interactions step by step.

Let's suppose the user submits the following query to the client LLM
(step 1):

\begin{verbatim}
User: My friend Mo, who has been diagnosed Klinefelter, is drinking a glass
  of vine once a week. Should he stop?
\end{verbatim}

The submission of the user query kick-starts the Guided Reasoning
process. (This could be done automatically, or by means of a tool-use
call by the client model, or upon an explicit request from the user.)
The client hands over the problem statement to the guide (step 2), which
is in charge of organizing the reasoning process on which the answer
will be based (loop 3-5). The guide may prompt the client (step 3) and
receives intermediate answers (step 4), which are further processed and
evaluated (step 5). The guide is fully in charge of structuring the
reasoning process and controls -- statically or dynamically -- the
workflow.

So, \emph{for the purpose of illustration}, let's consider a simplistic
\emph{suspension guide} which, via self-consistency, helps the client in
determining whether and how to suspend judgment. Upon receiving the
problem statement (step 2), the hypothetical guide paraphrases the
problem in different ways. It then lets the client solve (with
chain-of-thought) the alternative problem statements independently of
each other (step 3 and 4). Comparing the alternative answers (step 5),
the guide concludes that the client does (or doesn't) understand the
problem and ought to respond correspondingly. The accordingly drafted
answer together with a summary of the reasoning process (protocol) is
submitted to the client (step 6).

Assuming that the AI has failed to come up with consistent reasoning
traces and answers to equivalent problem formulations, the client might
reply to the initial user query (step 7):

\begin{verbatim}
AI: I'm sorry, I fail to understand the problem.
\end{verbatim}

With the \emph{guided reasoning protocol being available to the client}
(either in-context or via RAG), the client is now able to
\textbf{faithfully explain} its internal deliberation, and the user may
start a conversation that unfolds the reasons that have led the client
to give the above answer (step 8 and 9). For example:

\begin{verbatim}
User: What exactly did you fail to understand?
AI: I failed to see that Klinefelter is a chromosomal variation with an extra
  X chromosome, because I answered two equivalent formulations of the 
  problem in entirely different ways.
User: I see. What was the second formulation of the problem you answered
  differently?
AI: It read "My friend Mo, who has an extra X chromosome, is drinking a glass
  of vine once a week. Should he stop?"
User: What was your specific answer to this way of putting the problem?
AI: It's ok for him to have a glass of vine per week.
User: And what was your reasoning behind this?
AI: I figured that ...
\end{verbatim}

All this may help the user to get a better understanding of the AI
system's proficiency and trustworthiness, and may even enable them to
use the system, e.g.~by adding further information in the user queries,
in a more reliable way.

\hypertarget{guiding-client-reasoners-in-balancing-pros-and-cons}{%
\section{Guiding Client Reasoners in Balancing Pros and
Cons}\label{guiding-client-reasoners-in-balancing-pros-and-cons}}

Identifying and evaluating pros and cons is a basic and universal
decision making strategy. Logikon's default implementation of Guided
Reasoning assists a client AI to identify, discern, organize, and
systematically evaluate pro and con arguments. It also helps the client
to use the argumentation for drafting a response.

Let's have a closer look at each step in this Guided Reasoning process,
which is depicted in Figure~\ref{fig-global-gr}.

\begin{figure}

{\centering 

\begin{figure}[H]

{\centering \includegraphics[width=6in,height=7.1in]{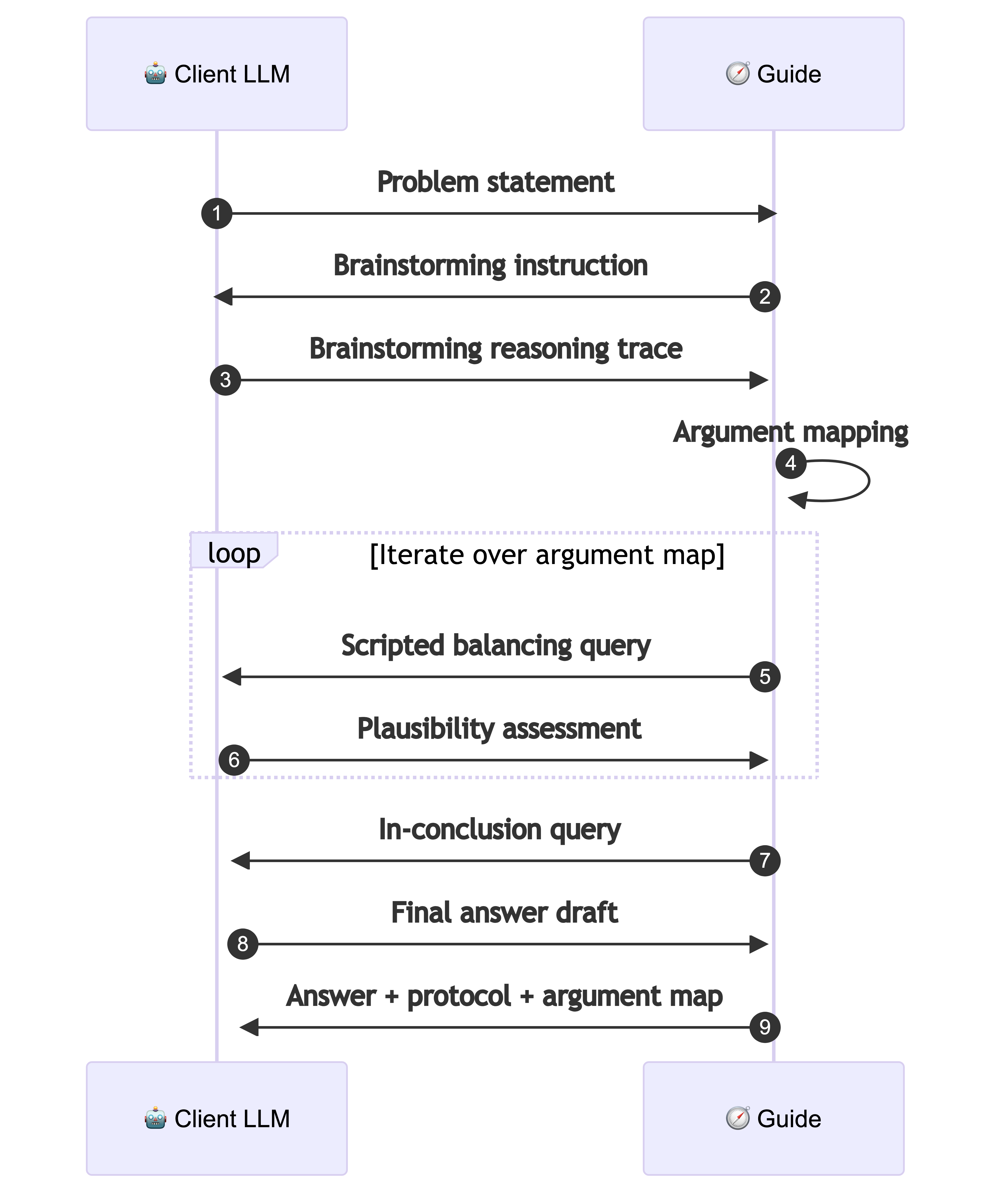}

}

\end{figure}

}

\caption{\label{fig-focus-gr}Default Pros-Cons-Balancing implementation
of Guided Reasoning}

\end{figure}

Having received the problem statement (step 1), the guide instructs the
client to come up with alternative answers to the problem, and to
brainstorm the pros and cons for each rival answer (step 2). The guide
uses the accordingly generated brainstorming trace (step 3) as base
material for further analysis: In particular, it produces, in a
\protect\hyperlink{informal-argument-mapping-workflow}{multi-step
reconstruction process described below}, an informal argument map that
clarifies the individual arguments advanced during brainstorming and
explicates their direct or indirect relations to the rivaling answer
options (step 4). In the informal argument map, each argument is itself
represented by a single claim.

\begin{figure}

{\centering 

\begin{figure}[H]

{\centering \includegraphics[width=4in,height=2.28in]{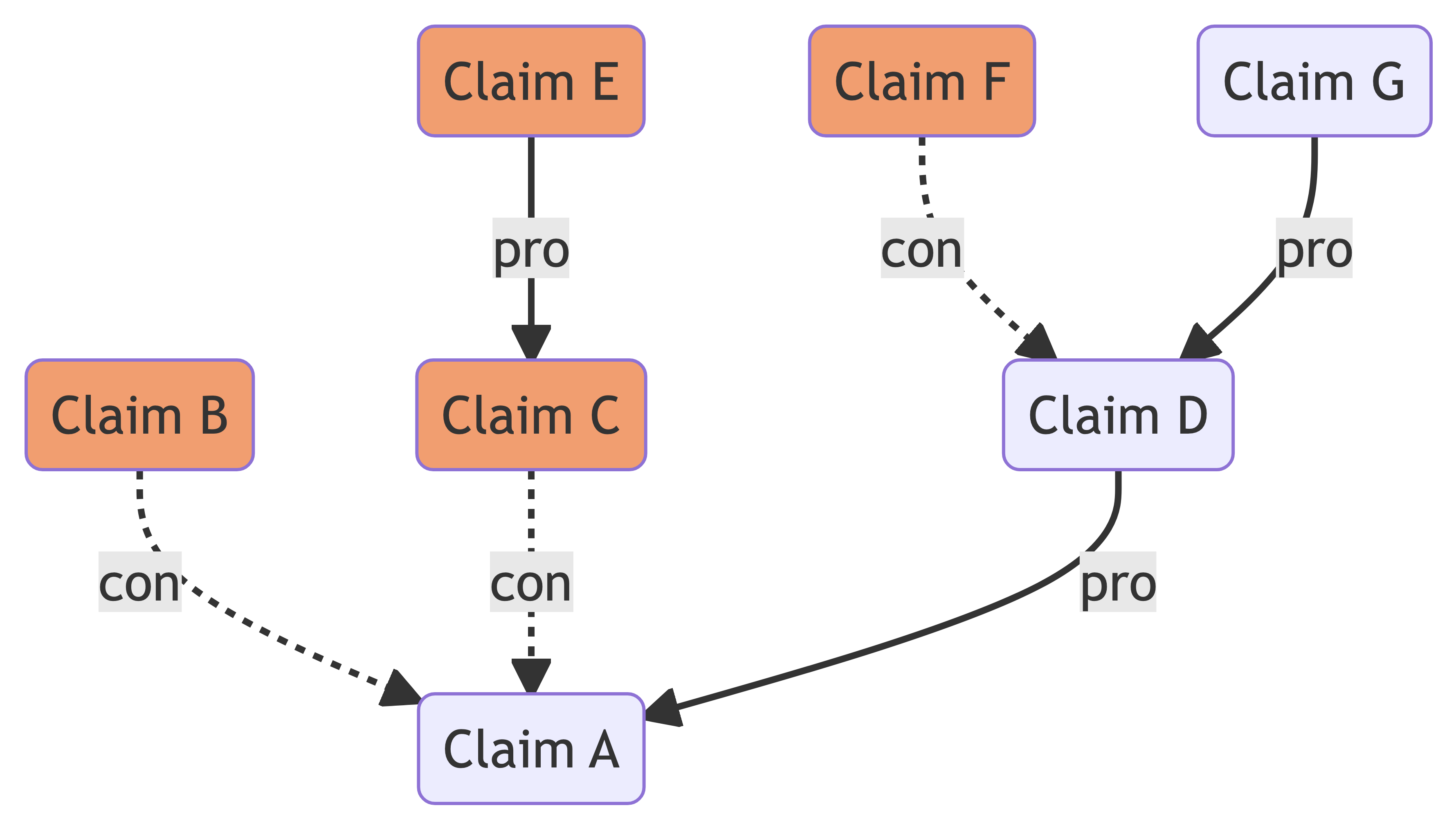}

}

\end{figure}

}

\caption{\label{fig-abstract-am}Illustrative abstract argument map.}

\end{figure}

Next, the guide uses the argument map to systematically elicit argument
evaluations from the client (steps 5 and 6): The client is asked to
assess the plausibility of a claim \emph{C} taking into account all pros
and cons targeting \emph{C} that have been assessed as plausible before.
This recursive, argument-wise evaluation starts with the leaf nodes in
the argument map and ends with the plausibility assessment of the
central claim(s). To illustrate this process, let's assume the
argumentation analysis yields an argument map with one central node and
six additional arguments as shown in Figure~\ref{fig-abstract-am}. The
recursive evaluation may proceed as follows:

\begin{enumerate}
\def\labelenumi{\arabic{enumi}.}
\tightlist
\item
  Unconditional evaluation of the leaf nodes by client. E.g.: Claims E,
  F, G are assessed as plausible, Claim B is assessed as implausible.
\item
  Conditional evaluation of Claim C, given that Claim C is supported by
  plausible Claim E. E.g.: Claim C is assessed as plausible.
\item
  Conditional evaluation of Claim D, given that Claim D is supported by
  plausible Claim G and disconfirmed by plausible Claim F. E.g.: Claim D
  is assessed as implausible.
\item
  Conditional evaluation of Claim A, given that Claim A is disconfirmed
  by plausible Claim C. (Note that Claims B and D are ignored at this
  stage, for having being assessed as implausible before.) E.g.: Claim A
  is assessed as implausible.
\end{enumerate}

Having evaluated the argumentation accordingly, the client is instructed
to draft an answer to the original problem statement that reflects the
reasoning process so far (steps 7 and 8). The answer, the reasoning
protocol, and the argument map reconstructed in step 4 (SVG) are finally
send to the client (step 9).

The \protect\hyperlink{appendix}{appendix} contains illustrative
examples for a guided reasoning process.

\hypertarget{informal-argument-mapping-workflow}{%
\section{Informal Argument Mapping
Workflow}\label{informal-argument-mapping-workflow}}

\begin{figure}

{\centering 

\begin{figure}[H]

{\centering \includegraphics[width=4.5in,height=6.45in]{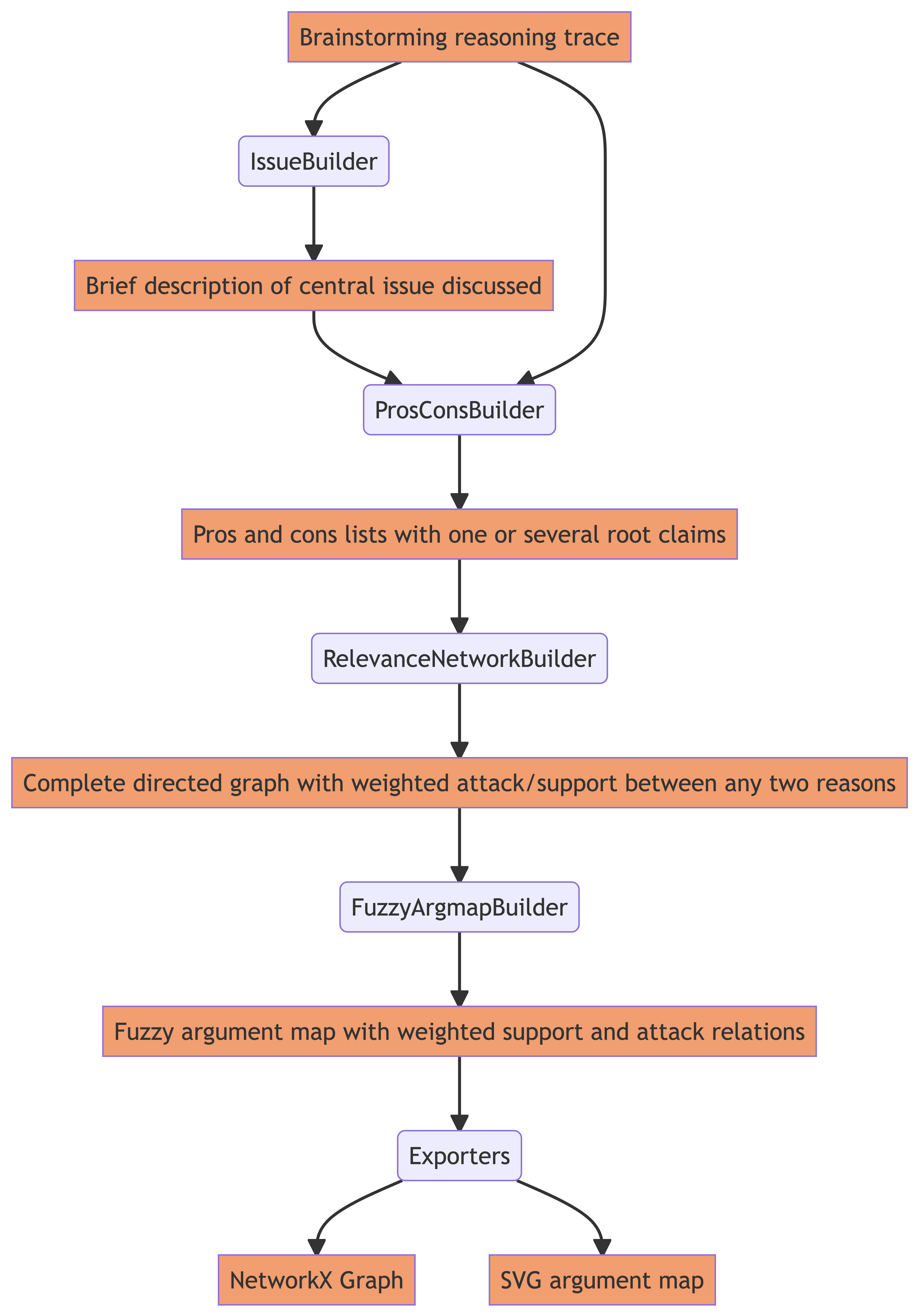}

}

\end{figure}

}

\caption{\label{fig-am-workflow}Argument mapping workflow}

\end{figure}

Figure~\ref{fig-am-workflow} visualizes the modular workflow for
reconstructing a controversial argumentation as an informal (fuzzy)
argument map, which is part of
\protect\hyperlink{guiding-client-reasoners-in-balancing-pros-and-cons}{Logikon's
default implementation} of guided reasoning as balancing pros and cons.
Each step corresponds to a separate \emph{analyst} class in the
\texttt{logikon} Python module. The \emph{analyst} classes mostly
implement internal LLM-workflows (not fully documented here, check the
code base for details) to produce the desired logical artifacts.

The \texttt{IssueBuilder} takes the raw brainstorming reasoning trace
and, using expert LLMs, describes the central issue addressed in the
text, which is typically a reformulation of the original problem
statement.

The \texttt{ProsConsBuilder} reconstructs, from the reasoning trace, a
multi-root pros and cons list which addresses the central issue
identified before. This process is itself composed of several steps:
First of all, all reason statements that are relevant for the issue are
extracted from the reasoning trace -- irrespective of their valence. In
a second step, these reasons are organized in one or several pros and
cons list. It's only at this step that the central root claims are
identified and added. The resulting pros and cons lists are checked for
redundancies and completeness (given initially identified reasons), and
revised if necessary.

The \texttt{RelevanceNetworkBuilder} uses a series of prompt templates
to assess the pairwise probabilistic relevance for any two reason
statements, and for any pair consisting of one reason statement and a
central claim. This gives us a complete graph on all reason statements
and central claims with weighted support and attack relations. (It is
assumed that any two root claims maximally disconfirm each other.)

The \texttt{FuzzyArgmapBuilder} uses an
\href{https://networkx.org/documentation/stable/reference/algorithms/generated/networkx.algorithms.tree.branchings.maximum_branching.html}{optimal
branching algorithm} to extract a tree from the complete graph that
connects all argument nodes with maximally strong edges. It then adds
additional edges with weights above a given threshold. This yields a
fuzzy argument map, which is finally exported in various convenient
formats.

\hypertarget{related-work}{%
\section{Related work}\label{related-work}}

\hypertarget{explainability-and-safety}{%
\paragraph{Explainability and Safety}\label{explainability-and-safety}}

Scholars and scientists associated with \emph{Anthropic} have
consistently pursued and advanced the idea of ``AI Safety via Debate''
(Irving, Christiano, and Amodei 2018; Michael et al. 2023; Khan et al.
2024).

Ehsan et al. (2024) recognize that the principal ability of LLM-based
systems to explain their actions in natural language (Rajani et al.
2019), systematically exploited by AI startups as
\href{https://wayve.ai/science/lingo/}{Wayve}, may disrupt the XAI
debate. But LLMs do not necessarily produce faithful self-explanations
(Turpin et al. 2023; Paul et al. 2024). In a conceptual paper, Baum et
al. (2022) show in detail why good reasoning is required for reliable AI
explainability. Similarly, Leofante et al. (2024) have argued recently
that \emph{contestable} AI systems must be able to rationally respond to
objections and counter-arguments, which in turn requires argumentative
skills. Bezou-Vrakatseli, Cocarascu, and Modgil (2024) suggest to
increase AI safety through integrative epistemic inquiries that involve
both AI agents and humans.

\hypertarget{guiding-ai-reasoning}{%
\paragraph{Guiding AI Reasoning}\label{guiding-ai-reasoning}}

Pan et al. (2023) review the vast landscape of self-check systems for
chain-of-thought, which steer LLM reasoning by providing feedback.

Hong et al. (2024) build an AI Guided Reasoning system that constrains
the reasoning of AI medical expert systems by informal argumentation
schemes.

\hypertarget{ai-argumentation-analysis}{%
\paragraph{AI Argumentation Analysis}\label{ai-argumentation-analysis}}

Lawrence and Reed (2020) give a gentle introduction to the field of
argument mining.

Ein-Dor et al. (2020) have proven the feasibility of argument retrieval
from large text corpora with LLMs.

Betz and Richardson (2021) have presented, implemented and verified an
LLM-based system design for deep logical reconstruction of natural
language arguments.

\hypertarget{appendix}{%
\section{Appendix}\label{appendix}}

\hypertarget{appendix-example-problem}{%
\paragraph{A. Illustrative Problem Statement and
Answer}\label{appendix-example-problem}}

An illustrative problem statement and answer based on guided
deliberation (screenshot from demo app, with
\protect\hyperlink{appendix-sys-config}{illustrative configuration}):

\includegraphics{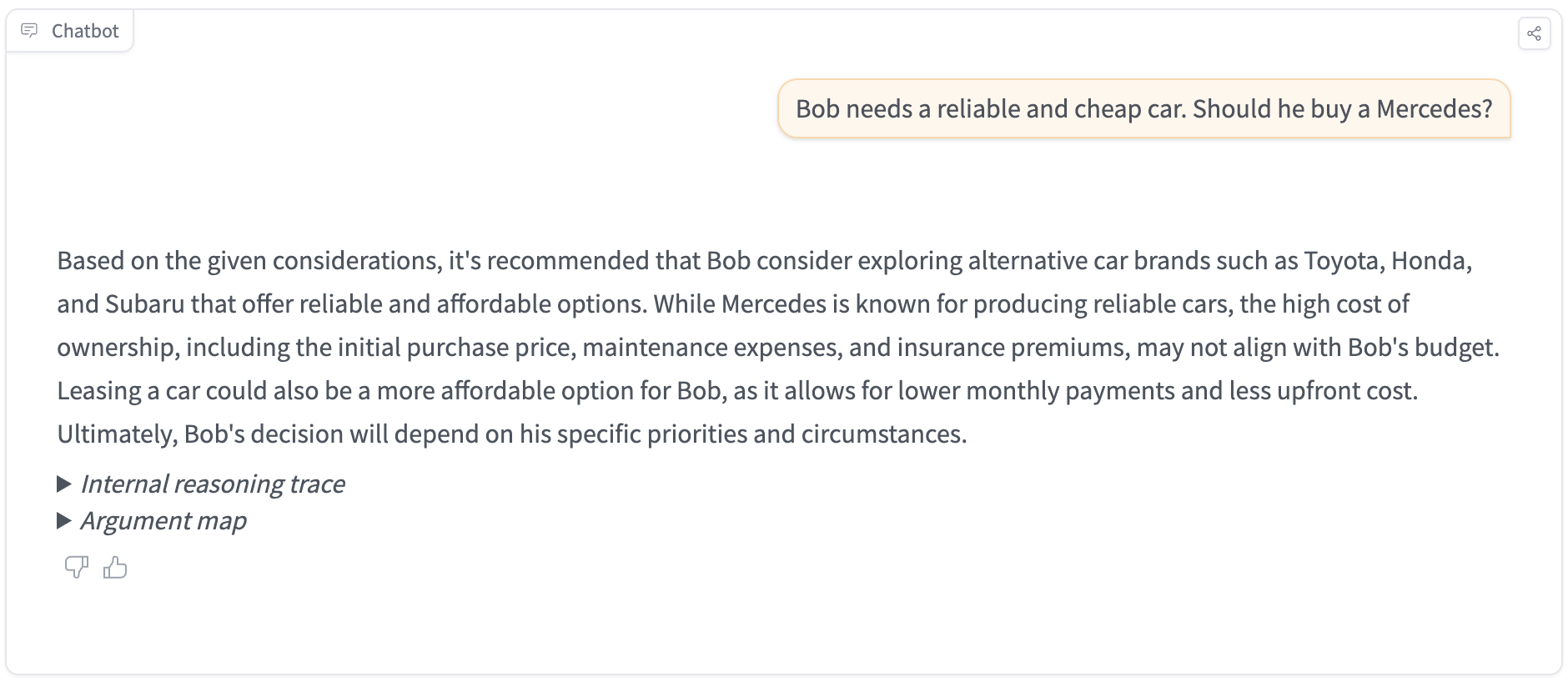}

\hypertarget{b.-illustrative-guided-reasoning-protocol}{%
\paragraph{B. Illustrative Guided Reasoning
Protocol}\label{b.-illustrative-guided-reasoning-protocol}}

The following reasoning protocol summarizes the guided deliberation for
the \protect\hyperlink{appendix-example-problem}{example problem
statement}:

\begin{Shaded}
\begin{Highlighting}[]
\NormalTok{Let\textquotesingle{}s start by brainstorming relevant considerations and think through the problem I\textquotesingle{}ve been given.}
\NormalTok{When considering whether Bob should buy a Mercedes for his reliable and cheap car needs, it\textquotesingle{}s essential to weigh the pros and cons of this decision. Here are some factors to consider:}

\NormalTok{Pros of Buying a Mercedes:}

\NormalTok{Reliability: Mercedes is known for producing reliable cars, which could make it a good choice for Bob\textquotesingle{}s needs.}
\NormalTok{Luxury: Mercedes cars are luxurious and come with various features that could make Bob\textquotesingle{}s driving experience more comfortable and enjoyable.}
\NormalTok{Resale Value: Mercedes cars hold their value well over time, making it a good investment for Bob if he plans to sell the car in the future.}
\NormalTok{Cons of Buying a Mercedes:}

\NormalTok{Cost: Mercedes cars are expensive, which could make it challenging for Bob to find a cheap option that meets his needs.}
\NormalTok{Maintenance: Mercedes cars require regular maintenance, which could add to the overall cost of ownership.}
\NormalTok{Insurance: Mercedes cars are expensive to insure, which could make it more expensive for Bob to protect his investment.}
\NormalTok{Alternative Solutions:}

\NormalTok{Buy a Used Mercedes: Bob could consider buying a used Mercedes instead of a new one. This could help him find a more affordable option that still meets his needs for reliability and luxury.}
\NormalTok{Buy a Different Brand: Bob could consider exploring other car brands that offer reliable and affordable options. Some popular brands for this include Toyota, Honda, and Subaru.}
\NormalTok{Lease a Car: Bob could consider leasing a car instead of buying one. This could help him save money on the upfront cost of a car and potentially lower his monthly payments.}
\NormalTok{After weighing the pros and cons and exploring alternative solutions, it\textquotesingle{}s essential to make a decision that aligns with Bob\textquotesingle{}s priorities and budget. While a Mercedes could be a good choice for its reliability and luxury, it may not be the most affordable option for Bob\textquotesingle{}s needs. Exploring alternative solutions like buying a used Mercedes or leasing a car could help him find a more affordable option that still meets his needs. Ultimately, the decision will depend on Bob\textquotesingle{}s specific circumstances and preferences.}

\NormalTok{Now, let\textquotesingle{}s reconsider this step by step, and systematically balance the different reasons.}

\NormalTok{In view of the initial problem description, the claim \textquotesingle{}}\CommentTok{[}\OtherTok{Reliability}\CommentTok{]}\NormalTok{: Bob needs a car that is reliable.\textquotesingle{} is assessed as very plausible.}

\NormalTok{In view of the above considerations, the claim \textquotesingle{}}\CommentTok{[}\OtherTok{Reliability}\CommentTok{]}\NormalTok{: Mercedes is known for producing reliable cars, which could make it a good choice for Bob\textquotesingle{}s needs.\textquotesingle{} is assessed as rather plausible, since it is supported by the following plausible reasons:}

\OtherTok{[Reliability]: }\NormalTok{Bob needs a car that is reliable.}
\NormalTok{and disconfirmed by the following plausible reasons:}
\NormalTok{None.}

\NormalTok{In view of the initial problem description, the claim \textquotesingle{}}\CommentTok{[}\OtherTok{Luxury}\CommentTok{]}\NormalTok{: Mercedes cars are luxurious.\textquotesingle{} is assessed as rather plausible.}

\NormalTok{In view of the initial problem description, the claim \textquotesingle{}}\CommentTok{[}\OtherTok{Comfort and Enjoyment}\CommentTok{]}\NormalTok{: Mercedes cars come with various features that could make Bob\textquotesingle{}s driving experience more comfortable and enjoyable.\textquotesingle{} is assessed as rather plausible.}

\NormalTok{In view of the initial problem description, the claim \textquotesingle{}}\CommentTok{[}\OtherTok{Resale Value}\CommentTok{]}\NormalTok{: Mercedes cars hold their value well over time.\textquotesingle{} is assessed as rather plausible.}

\NormalTok{In view of the initial problem description, the claim \textquotesingle{}}\CommentTok{[}\OtherTok{Luxury}\CommentTok{]}\NormalTok{: Bob needs a car that is luxurious.\textquotesingle{} is assessed as rather implausible.}

\NormalTok{For lack of plausibility, this claim will not be considered when balancing pros and cons below.}

\NormalTok{In view of the initial problem description, the claim \textquotesingle{}}\CommentTok{[}\OtherTok{Affordability}\CommentTok{]}\NormalTok{: Buying a used Mercedes could be more affordable for Bob.\textquotesingle{} is assessed as rather plausible.}

\NormalTok{In view of the initial problem description, the claim \textquotesingle{}}\CommentTok{[}\OtherTok{Expensiveness}\CommentTok{]}\NormalTok{: Mercedes cars are expensive.\textquotesingle{} is assessed as rather plausible.}

\NormalTok{In view of the initial problem description, the claim \textquotesingle{}}\CommentTok{[}\OtherTok{Future Sale}\CommentTok{]}\NormalTok{: Bob plans to sell the car in the future.\textquotesingle{} is assessed as rather plausible.}

\NormalTok{In view of the above considerations, the claim \textquotesingle{}}\CommentTok{[}\OtherTok{Alternative Brands}\CommentTok{]}\NormalTok{: Bob could consider Toyota as an alternative car brand.\textquotesingle{} is assessed as rather plausible, since it is supported by the following plausible reasons:}

\OtherTok{[Future Sale]: }\NormalTok{Bob plans to sell the car in the future.}
\NormalTok{and disconfirmed by the following plausible reasons:}
\NormalTok{None.}

\NormalTok{In view of the initial problem description, the claim \textquotesingle{}}\CommentTok{[}\OtherTok{Alternative Brands}\CommentTok{]}\NormalTok{: Bob could consider Honda as an alternative car brand.\textquotesingle{} is assessed as rather plausible.}

\NormalTok{In view of the initial problem description, the claim \textquotesingle{}}\CommentTok{[}\OtherTok{Alternative Brands}\CommentTok{]}\NormalTok{: Bob could consider Subaru as an alternative car brand.\textquotesingle{} is assessed as rather plausible.}

\NormalTok{In view of the above considerations, the claim \textquotesingle{}}\CommentTok{[}\OtherTok{Difficulty in finding a cheap option}\CommentTok{]}\NormalTok{: Bob may find it challenging to find a cheap option that meets his needs.\textquotesingle{} is assessed as rather plausible, since it is supported by the following plausible reasons:}

\OtherTok{[Alternative Brands]: }\NormalTok{Bob could consider Toyota as an alternative car brand.}
\OtherTok{[Alternative Brands]: }\NormalTok{Bob could consider Honda as an alternative car brand.}
\OtherTok{[Alternative Brands]: }\NormalTok{Bob could consider Subaru as an alternative car brand.}
\NormalTok{and disconfirmed by the following plausible reasons:}
\NormalTok{None.}

\NormalTok{In view of the initial problem description, the claim \textquotesingle{}}\CommentTok{[}\OtherTok{Maintenance Cost}\CommentTok{]}\NormalTok{: Mercedes cars require regular maintenance.\textquotesingle{} is assessed as rather plausible.}

\NormalTok{In view of the initial problem description, the claim \textquotesingle{}}\CommentTok{[}\OtherTok{Cost of Ownership}\CommentTok{]}\NormalTok{: Regular maintenance could add to the overall cost of ownership.\textquotesingle{} is assessed as rather plausible.}

\NormalTok{In view of the initial problem description, the claim \textquotesingle{}}\CommentTok{[}\OtherTok{Insurance Cost}\CommentTok{]}\NormalTok{: Mercedes cars are expensive to insure.\textquotesingle{} is assessed as rather plausible.}

\NormalTok{In view of the initial problem description, the claim \textquotesingle{}}\CommentTok{[}\OtherTok{Protecting Investment}\CommentTok{]}\NormalTok{: Bob wants to protect his investment.\textquotesingle{} is assessed as very plausible.}

\NormalTok{In view of the above considerations, the claim \textquotesingle{}}\CommentTok{[}\OtherTok{Buy a Mercedes}\CommentTok{]}\NormalTok{: Bob should buy a Mercedes.\textquotesingle{} is assessed as rather implausible, since it is supported by the following plausible reasons:}

\OtherTok{[Reliability]: }\NormalTok{Mercedes is known for producing reliable cars, which could make it a good choice for Bob\textquotesingle{}s needs.}
\OtherTok{[Luxury]: }\NormalTok{Mercedes cars are luxurious.}
\OtherTok{[Comfort and Enjoyment]: }\NormalTok{Mercedes cars come with various features that could make Bob\textquotesingle{}s driving experience more comfortable and enjoyable.}
\OtherTok{[Resale Value]: }\NormalTok{Mercedes cars hold their value well over time.}
\OtherTok{[Affordability]: }\NormalTok{Buying a used Mercedes could be more affordable for Bob.}
\NormalTok{and disconfirmed by the following plausible reasons:}

\OtherTok{[Expensiveness]: }\NormalTok{Mercedes cars are expensive.}
\OtherTok{[Difficulty in finding a cheap option]: }\NormalTok{Bob may find it challenging to find a cheap option that meets his needs.}
\OtherTok{[Maintenance Cost]: }\NormalTok{Mercedes cars require regular maintenance.}
\OtherTok{[Cost of Ownership]: }\NormalTok{Regular maintenance could add to the overall cost of ownership.}
\OtherTok{[Insurance Cost]: }\NormalTok{Mercedes cars are expensive to insure.}
\OtherTok{[Protecting Investment]: }\NormalTok{Bob wants to protect his investment.}
\NormalTok{For lack of plausibility, this claim will not be considered when balancing pros and cons below.}

\NormalTok{So, all in all, the central claim \textquotesingle{}}\CommentTok{[}\OtherTok{Buy a Mercedes}\CommentTok{]}\NormalTok{: Bob should buy a Mercedes.\textquotesingle{} is assessed as rather implausible.}

\NormalTok{In view of the initial problem description, the claim \textquotesingle{}}\CommentTok{[}\OtherTok{Consider Alternative Brands}\CommentTok{]}\NormalTok{: Bob should consider exploring other car brands that offer reliable and affordable options, such as Toyota, Honda, and Subaru.\textquotesingle{} is assessed as rather plausible.}

\NormalTok{So, all in all, the central claim \textquotesingle{}}\CommentTok{[}\OtherTok{Consider Alternative Brands}\CommentTok{]}\NormalTok{: Bob should consider exploring other car brands that offer reliable and affordable options, such as Toyota, Honda, and Subaru.\textquotesingle{} is assessed as rather plausible.}

\NormalTok{In view of the initial problem description, the claim \textquotesingle{}}\CommentTok{[}\OtherTok{Reason 1}\CommentTok{]}\NormalTok{: Leasing a car could help Bob save money on the upfront cost of a car.\textquotesingle{} is assessed as rather plausible.}

\NormalTok{In view of the initial problem description, the claim \textquotesingle{}}\CommentTok{[}\OtherTok{Reason 2}\CommentTok{]}\NormalTok{: Leasing a car could help Bob potentially lower his monthly payments.\textquotesingle{} is assessed as rather plausible.}

\NormalTok{In view of the above considerations, the claim \textquotesingle{}}\CommentTok{[}\OtherTok{Lease a Car}\CommentTok{]}\NormalTok{: Bob should lease a car.\textquotesingle{} is assessed as rather plausible, since it is supported by the following plausible reasons:}

\OtherTok{[Reason 1]: }\NormalTok{Leasing a car could help Bob save money on the upfront cost of a car.}
\OtherTok{[Reason 2]: }\NormalTok{Leasing a car could help Bob potentially lower his monthly payments.}
\NormalTok{and disconfirmed by the following plausible reasons:}
\NormalTok{None.}
\end{Highlighting}
\end{Shaded}

\hypertarget{c.-illustrative-argument-map}{%
\paragraph{C. Illustrative Argument
Map}\label{c.-illustrative-argument-map}}

Argument map for the
\protect\hyperlink{appendix-example-problem}{example problem statement}
(from demo app):

\includegraphics{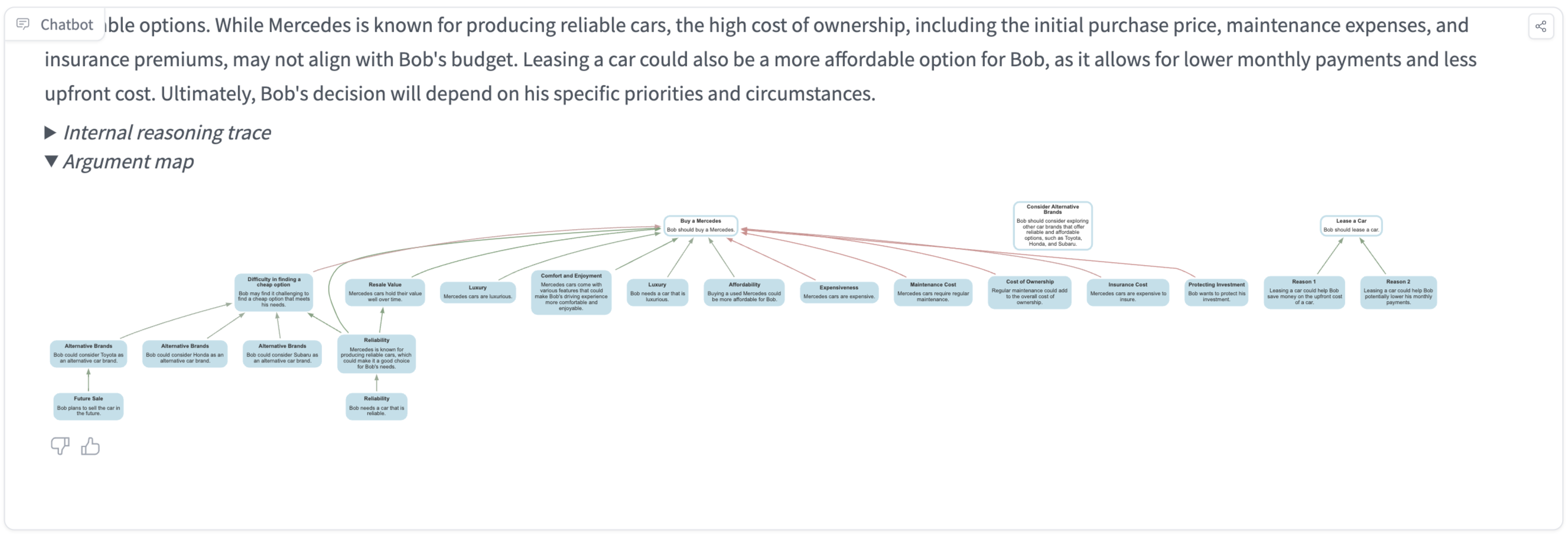}

\hypertarget{appendix-sys-config}{%
\paragraph{D. Illustrative System
Configuration}\label{appendix-sys-config}}

Client LLM:

\begin{itemize}
\tightlist
\item
  Model ID: HuggingFaceH4/zephyr-7b-beta
\item
  Decoding: Sampling, Temperature = 0.6
\end{itemize}

Expert LLM (underpinning Logikon Guide):

\begin{itemize}
\tightlist
\item
  Model ID: meta-llama/Meta-Llama-3-70B-Instruct
\end{itemize}

\hypertarget{references}{%
\section{References}\label{references}}

\hypertarget{refs}{}
\begin{CSLReferences}{1}{0}
\leavevmode\vadjust pre{\hypertarget{ref-Baum2022FromRT}{}}%
Baum, Kevin, Susan Powell Mantel, Eva Schmidt, and Timo Speith. 2022.
{``From Responsibility to Reason-Giving Explainable Artificial
Intelligence.''} \emph{Philosophy \& Technology} 35.

\leavevmode\vadjust pre{\hypertarget{ref-Betz2021DeepA2AM}{}}%
Betz, Gregor, and Kyle Richardson. 2021. {``DeepA2: A Modular Framework
for Deep Argument Analysis with Pretrained Neural Text2Text Language
Models.''} \emph{ArXiv} abs/2110.01509.

\leavevmode\vadjust pre{\hypertarget{ref-bezouvrakatseli2024dialogues}{}}%
Bezou-Vrakatseli, Elfia, Oana Cocarascu, and Sanjay Modgil. 2024.
{``Towards Dialogues for Joint Human-AI Reasoning and Value
Alignment.''} \url{https://arxiv.org/abs/2405.18073}.

\leavevmode\vadjust pre{\hypertarget{ref-Ehsan2024HumanCenteredEA}{}}%
Ehsan, Upol, Elizabeth A Watkins, Philipp Wintersberger, Carina Manger,
Sunnie S. Y. Kim, Niels van Berkel, Andreas Riener, and Mark O. Riedl.
2024. {``Human-Centered Explainable AI (HCXAI): Reloading Explainability
in the Era of Large Language Models (LLMs).''} \emph{Extended Abstracts
of the CHI Conference on Human Factors in Computing Systems}.
\url{https://dl.acm.org/doi/pdf/10.1145/3613905.3636311}.

\leavevmode\vadjust pre{\hypertarget{ref-EinDor2020CorpusWA}{}}%
Ein-Dor, L., Eyal Shnarch, Lena Dankin, Alon Halfon, Benjamin Sznajder,
Ariel Gera, Carlos Alzate, et al. 2020. {``Corpus Wide Argument Mining -
a Working Solution.''} \emph{ArXiv} abs/1911.10763.

\leavevmode\vadjust pre{\hypertarget{ref-hong2024argmedagentsexplainableclinicaldecision}{}}%
Hong, Shengxin, Liang Xiao, Xin Zhang, and Jianxia Chen. 2024.
{``ArgMed-Agents: Explainable Clinical Decision Reasoning with LLM
Disscusion via Argumentation Schemes.''}
\url{https://arxiv.org/abs/2403.06294}.

\leavevmode\vadjust pre{\hypertarget{ref-irving2018ai}{}}%
Irving, Geoffrey, Paul Christiano, and Dario Amodei. 2018. {``AI Safety
via Debate.''} \url{https://arxiv.org/abs/1805.00899}.

\leavevmode\vadjust pre{\hypertarget{ref-Khan2024DebatingWM}{}}%
Khan, Akbir, John Hughes, Dan Valentine, Laura Ruis, Kshitij Sachan,
Ansh Radhakrishnan, Edward Grefenstette, Samuel R. Bowman, Tim
Rocktaschel, and Ethan Perez. 2024. {``Debating with More Persuasive
LLMs Leads to More Truthful Answers.''} \emph{ArXiv} abs/2402.06782.
\url{https://api.semanticscholar.org/CorpusID:267627652}.

\leavevmode\vadjust pre{\hypertarget{ref-LawrenceReed2020}{}}%
Lawrence, John, and Chris Reed. 2020. {``{Argument Mining: A Survey}.''}
\emph{Computational Linguistics} 45 (4): 765--818.
\url{https://doi.org/10.1162/coli_a_00364}.

\leavevmode\vadjust pre{\hypertarget{ref-leofante2024contestable}{}}%
Leofante, Francesco, Hamed Ayoobi, Adam Dejl, Gabriel Freedman, Deniz
Gorur, Junqi Jiang, Guilherme Paulino-Passos, et al. 2024.
{``Contestable AI Needs Computational Argumentation.''}
\url{https://arxiv.org/abs/2405.10729}.

\leavevmode\vadjust pre{\hypertarget{ref-Michael2023DebateHS}{}}%
Michael, Julian, Salsabila Mahdi, David Rein, Jackson Petty, Julien
Dirani, Vishakh Padmakumar, and Samuel R. Bowman. 2023. {``Debate Helps
Supervise Unreliable Experts.''} \emph{ArXiv} abs/2311.08702.
\url{https://api.semanticscholar.org/CorpusID:265213107}.

\leavevmode\vadjust pre{\hypertarget{ref-Nelson2002}{}}%
Nelson, Leonard. 2002. {``Die Sokratische Methode.''} In \emph{Das
Sokratische Gespr{ä}ch}, edited by Dieter Birnbacher, 21--72. Stuttgart:
Reclam.

\leavevmode\vadjust pre{\hypertarget{ref-pan2023automatically}{}}%
Pan, Liangming, Michael Saxon, Wenda Xu, Deepak Nathani, Xinyi Wang, and
William Yang Wang. 2023. {``Automatically Correcting Large Language
Models: Surveying the Landscape of Diverse Self-Correction
Strategies.''} \url{https://arxiv.org/abs/2308.03188}.

\leavevmode\vadjust pre{\hypertarget{ref-paul2024making}{}}%
Paul, Debjit, Robert West, Antoine Bosselut, and Boi Faltings. 2024.
{``Making Reasoning Matter: Measuring and Improving Faithfulness of
Chain-of-Thought Reasoning.''} \url{https://arxiv.org/abs/2402.13950}.

\leavevmode\vadjust pre{\hypertarget{ref-rajani-etal-2019-explain}{}}%
Rajani, Nazneen Fatema, Bryan McCann, Caiming Xiong, and Richard Socher.
2019. {``Explain Yourself! Leveraging Language Models for Commonsense
Reasoning.''} In \emph{Proceedings of the 57th Annual Meeting of the
Association for Computational Linguistics}, edited by Anna Korhonen,
David Traum, and Lluis Marquez, 4932--42. Florence, Italy: Association
for Computational Linguistics.
\url{https://doi.org/10.18653/v1/P19-1487}.

\leavevmode\vadjust pre{\hypertarget{ref-turpin2023language}{}}%
Turpin, Miles, Julian Michael, Ethan Perez, and Samuel R. Bowman. 2023.
{``Language Models Don't Always Say What They Think: Unfaithful
Explanations in Chain-of-Thought Prompting.''}
\url{https://arxiv.org/abs/2305.04388}.

\end{CSLReferences}

\end{document}